\documentclass[letterpaper]{article}
\usepackage{aaai}
\usepackage{times}
\usepackage{url}
\usepackage{helvet}
\usepackage{courier}
\usepackage{graphicx}
\usepackage{amsmath,amssymb}
\usepackage{booktabs}
\usepackage[dvipsnames]{xcolor}
\begin{document}
%
\title{Clean \& Clear: Feasibility of Safe LLM Clinical Guidance}
\author{\textbf{Julia Ive\textsuperscript{1}}, \textbf{Felix Jozsa\textsuperscript{2,4}}, \textbf{Nick Jackson\textsuperscript{3}},
\textbf{Paulina Bondaronek\textsuperscript{1}},
\textbf{Ciaran Scott Hill\textsuperscript{2}},
\textbf{Richard Dobson\textsuperscript{1,5}} \\
\textsuperscript{1} {University College London, London, UK}\\
\textsuperscript{2} {Wolfson Institute of Biomedical Research, University College London, UK}\\
\textsuperscript{3} {King's College Hospital, Denmark Hill, London, UK} \\
\textsuperscript{4} {National Hospital for Neurology and Neurosurgery, Queen Square, London, UK} \\
\textsuperscript{5} {King's College London, London, UK}}
\maketitle

\begin{abstract}
\begin{quote}

\textbf{Background:}

Clinical guidelines are central to safe evidence-based medicine in modern healthcare, providing diagnostic criteria, treatment options and monitoring advice for a wide range of illnesses that a clinician may encounter in the community or hospital setting. Local guidelines are typically stored on a repository in the hospital intranet, but finding the correct guideline, before reading and extracting the correct clinical information is time consuming for clinicians.

LLM-empowered chatbots have shown great promise in Healthcare Q\&A tasks, offering the potential to provide quick and accurate responses to medical inquiries. However, despite their potential, there are significant concerns regarding their reliability and the occurrence of hallucinations—instances where the chatbot generates incorrect or nonsensical information. These issues can impede their effective integration into healthcare settings

\textbf{Objectives:}

Our main objectives are: (1) the development of LLM-empowered chatbot software capable of reliably answering clinical guideline questions using University College London Hospital (UCLH) clinical guidelines; and (2) to conduct a preliminary assessment of this tool's utility in clinical practice by evaluating the trade-off between the quality of extracted clinical information and the retrieval time when deploying it in simulated clinical situations.

\textbf{Methods:}
We used the open-weight Llama-3.1-8B LLM to extract relevant information from the UCLH guidelines to answer questions. Our approach highlights the safety and reliability of referencing information over its interpretation and response generation, ensuring adherence to established guidelines. Seven doctors from the ward assessed the chatbot's performance by comparing its answers to the gold standard, evaluating relevance and completeness, and measuring the time taken to find answers manually versus using the chatbot.

\textbf{Results:}
Our chatbot demonstrates promising performance in terms of relevance, with $\sim$73\% of its responses rated as very relevant, showcasing a strong understanding of the clinical context. Importantly, our chatbot achieves a recall of 100\% for extracted guideline lines, substantially minimising the risk of missing critical information. Approximately 78\% of responses were rated satisfactory in terms of completeness. A small portion ($\sim$14.5\%) contained minor unnecessary information, indicating occasional lapses in precision.  The chatbot' showed high efficiency, with an average completion time of 10 seconds, compared to 30 seconds for human respondents. Evaluation of clinical reasoning showed that 72\% of the chatbot's responses were without flaws.

\textbf{Conclusions:}
Our chatbot demonstrates significant potential to speed up and improve the process of accessing locally relevant clinical information for healthcare professionals by providing quick, reliable answers based on clinical guidelines. 

\end{quote}
\end{abstract}

%


\section{Introduction}

Clinical guidelines are consensus documents that use an evidence-based approach to compile clear instructions for healthcare professionals on how to assess, treat, and escalate care for patients with a particular illness or condition \cite{guerra2023clinical}. The General Medical Council (GMC) in the UK enshrines the requirement for doctors to formulate treatment plans ``according to established principles and best evidence''\footnote{\url{https://www.gmc-uk.org/education/standards-guidance-and-curricula/standards-and-outcomes/outcomes-for-graduates}} and clearly describes the use of clinical guidelines as being part of ``good medical practice'' \cite{Scraggs2012}. Furthermore,  evidence from multiple international studies demonstrates improved patient safety outcomes associated with clinician adherence to clinical guidelines \cite{Kresevic2024-se,dean2001decreased,vanwagner2020blood,arnold2009improving,monti2023effect,coba2011resuscitation}.

Whilst the National Institute for Health and Care Excellence (NICE) provides UK-wide clinical guidelines\footnote{\url{https://www.nice.org.uk/guidance}}, hospitals also formulate local clinical guidelines, and these are the main resource used by doctors when needing clinical information to treat patients. These local guidelines are often broadly based around national NICE guidelines, but have region-specific information relating to the local patient population (for example, particular recommendations for first-line antibiotic treatment of urinary infections guided by local pathogens and sensitivities \cite{curtis2018time}). Furthermore, local hospital guidelines provide contact details for relevant team members if further discussion or escalation is required by the clinician, information which cannot be provided by the NICE guidelines. Thus, typically a clinician will consult the hospital clinical guideline relevant to the patient condition they are treating, requiring specific discrete answers to relevant questions such as, for example, ``What antibiotics should I prescribe for meningitis'', or ``How should I monitor a patient with hyperkalaemia?''

Despite widespread acceptance of clinical guidelines as the gold standard tool in clinical practice, there are barriers to their implementation in daily patient care. Multiple studies have identified that guideline complexity \cite{qumseya2021barriers} and time constraints \cite{tsiga2013influence,freedman2021docs,van2020searching} are key factors that impede clinician adherence to guideline use. Given that today's healthcare environment involves both increased time pressures for clinicians \cite{torjesen2021doctors,daniels2024perceived}, and increasing guideline complexity \cite{kann2020changes}, implementation of an Artificial Intelligence (AI) empowered tool to aid with this process could be significantly beneficial and ultimately improve patient care.

Large Language Model (LLM) empowered chatbots have shown great promise in healthcare Q\&A tasks \cite{park2024assessing}, offering the potential to provide quick and accurate responses to medical inquiries \cite{Kresevic2024-se,williams2024evaluating}. However, despite their potential, there are significant concerns regarding their reliability and the occurrence of hallucinations, instances where the chatbot generates incorrect or non-sensical information~\cite{mesko-topol-23}. These issues impede their effective integration into healthcare settings, because the generation of incorrect information could lead to adverse events and patient harm in a clinical setting. 

To address these limitations, we focus on a safer approach to leverage LLMs for evidence extraction rather than generation. By focusing on extracting text-based insights for clinical interpretation, this method reduces the risks associated with hallucinations and inaccuracies while preserving the benefits of automated data processing. In healthcare applications, smaller models like FLAN-T5 have shown promising results in zero-shot evidence extraction, identifying cases such as postpartum hemorrhage more effectively than traditional methods~\cite{Alsentzer2023}. In the general domain, larger models have recently showed particular promise in extracting information from documents with rich layout \cite{Colakoglu2025-qt}.

\paragraph{Study Aims and Objectives}

The aim of this study was to design an LLM chatbot capable of answering clinical queries confined to a local set of clinical guidelines and implement it in a simulated hospital setting with doctors to preliminary assess reliability and clinical utility of its answers.

Our primary objectives were twofold:

\begin{itemize}
    \item Leverage LLM information extraction capacities to create advanced chatbot software to reliably answer questions based on University College London Hospital (UCLH) clinical guidelines. 
    \item A preliminary assessment of the trade-offs involved in deploying this chatbot in the real-world setting, particularly focusing on the balance between the quality of the extracted clinical information and the retrieval time.
\end{itemize}

\section{Methods}

\paragraph{Ethical Considerations }

This study was approved by the UCL Review Board as TRE-467283, and approved as a clinical service audit at the National Hospital for Neurology and Neurosurgery, UCLH, London as 130-202425-SE.

\paragraph{LLM Adaptation}

In our study, we used the open-weight Llama-3.1-8B LLM~\footnote{\url{https://huggingface.co/meta-llama/Llama-3.1-8B-Instruct}} \cite{llama3}. Llama-3 is one of the most popular non-commercial models. This model was chosen for its ability to be deployed locally, ensuring the secure handling of private clinical data and adherence to computational constraints. We preserved the default inference hyperparameters on an A10G Tensor Core GPU 24 Gb unit.

The model was used in a few-shot setting, using two examples, and the process was divided into the following two steps: 

\begin{enumerate}
    \item Topic identification: determine the topic of the given user question.
    \item Line identification: identify the minimum relevant lines from the guideline context that help answer the question.
\end{enumerate}

Examples of our prompts and sample answers are detailed in Table~\ref{tab:summary_answers}. Regarding the prompts, both are designed to ensure outputs in a specific format that can be easily post-processed. These outputs are either a topic from a pre-set list or line numbers from the input guideline. This pre-set format guarantees output stability, even considering the variability of user questions. The stability of these outputs was confirmed during our experiments at the development stage.

\begin{table*}[h]
    \centering
    \scalebox{0.9}{
    \begin{tabular}{p{2cm}p{16cm}}
    \toprule
    \textbf{Step} & \textbf{Example Prompt} \\
        \toprule
         Topic identification & \textbf{User}: Identify the topic of the following question: \newline How do I define hypertensive emergency? \newline Use the following topic list: \newline Hypertension, Diabetes, Heart Disease. \newline Do not include any other text. Provide only the topic name. \newline 
         \textbf{Assistant}: Hypertension \\
        \midrule
       Line identification & \textbf{User}:Your question is: \newline How do I define hypertensive emergency? \newline Identify the minimum relevant lines from the context below that help answer the question. \newline Do not include any other text. Provide the line numbers that answer the question, separated by white space. \newline The content of the sections is below: \newline 0: DEFINITIONS, SYMPTOMS \& CAUSES: Hypertensive Emergency, also known as Accelerated Hypertension, is characterized by severe hypertension ($>$180/120 mmHg) accompanied by new or worsening target organ damage. \newline
       1: DEFINITIONS, SYMPTOMS \& CAUSES: Hypertensive Urgency refers to severe hypertension without target organ damage.\newline
       2: DEFINITIONS, SYMPTOMS \& CAUSES: Symptoms can include headache, visual disturbances, chest pain, shortness of breath, dizziness, and neurological deficits. \newline
       3: CAUSES (not exhaustive): Uncontrolled Primary Hypertension \newline
       4: CAUSES (not exhaustive): Secondary Hypertension (e.g., renal artery stenosis, primary hyperaldosteronism, pheochromocytoma)...  \newline
       \textbf{Assistant}: 1 \\
        \bottomrule
    \end{tabular}}
    \caption{\textbf{Example Prompts for Topic and Line Identification for our Guideline Q\&A.} Our system first classifies the user’s query under a predefined topic list (e.g., Hypertension, Diabetes, Heart Disease). Next, it extracts the minimum relevant lines from a provided clinical context. All the examples have been paraphrased.}
    \label{tab:summary_answers}
\end{table*}

Referencing only the relevant lines from a guideline is crucial for ensuring the safety and relevance of the information provided. By focusing on the information selection, we can significantly reduce the risk of hallucinations, inaccuracies and biases. 

However, providing relevant line numbers is insufficient; we need to understand the rationale behind the model's decisions. To address this, we investigated the reasoning capacities of Llama by examining the outputs of its DeepSeek version \texttt{DeepSeek-R1-Distill-Llama-8B} \cite{deepseekai2025reasoningcapability}. This DeepSeek model was distilled from Llama and then trained via large-scale reinforcement learning (RL) to develop enhanced reasoning capacities. We presented outputs of DeepSeek-Llama to human clinical experts for their evaluation.

\section{Evaluation Design}

The assessment design for this study involved both automatic and human evaluation. Automatic evaluation focused on the effectiveness of line extraction per question. Three key metrics were used: precision, recall, and F-score. Precision measures the accuracy of the extracted lines, indicating the proportion of relevant lines among those extracted. Recall assesses the completeness, showing the proportion of relevant lines that were successfully extracted from the total relevant lines available. The F-score combines precision and recall into a single metric, providing a balanced measure. 

Human evaluation included topic-based and scenario-based evaluations by human in-domain experts in an actual hospital environment at the National Hospital for Neurology and Neurosurgery (NHNN), a tertiary neurosciences centre.

\subsection{Participants}

Seven assessors were recruited to evaluate our chatbot. All were medical doctors with between 1 and 4 years postgraduate clinical experience. Six participants were junior doctors in the neurosurgery department of NHNN, and one assessor was a General Practice trainee. 

Six guidelines were randomly chosen from the UCLH general medical guidelines and screened to ensure that they contained similar amounts of textual information and were representative of the format and size of the other UCLH guidelines. The chosen guidelines were hot swollen joint, pancreatitis, pericarditis, severe hypertension, hypokalaemia, and diabetic ketoacidosis (DKA).
Each assessor evaluated nine chatbot answers across three different guidelines from the selected group of six. For each guideline, three questions were asked: two topics featured predefined questions, while one topic allowed experts to choose their own questions (see Table \ref{tab:task_questions}). The topics and questions for the predefined setup, along with their corresponding gold standard answers, were selected by the authors (FJ and NJ). These data were also used in the automatic evaluation setup.

\begin{figure}[ht]
\centering
\includegraphics[width=.9\linewidth]{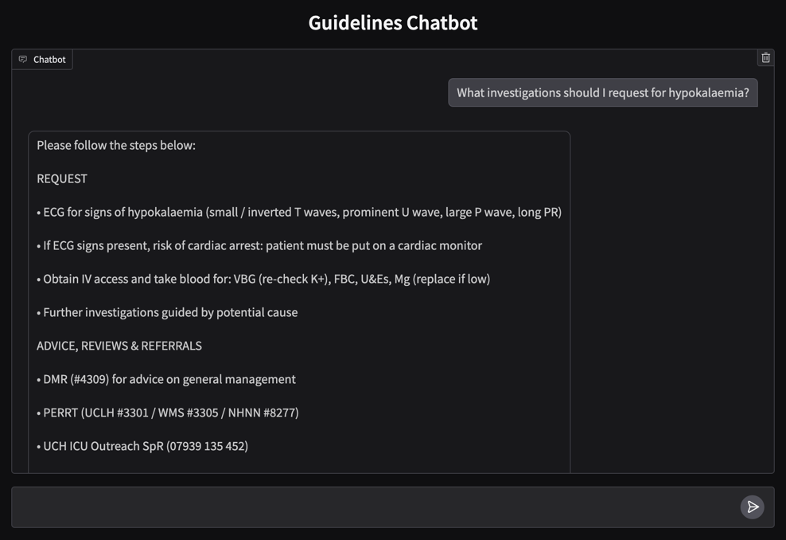}
\caption{\textbf{Our Chatbot Providing Investigation Recommendations for Hypokalaemia.} This screenshot displays an interaction with our chatbot. } 
\label{fig:interface}
\end{figure}

The assessment was conducted during in-person sessions using a computer with dedicated access to the chatbot interface within the Trusted Research Environment (TRE) provided by UCL specifically for testing purposes (see Figure \ref{fig:interface}). This setup ensured a secure and controlled environment for evaluation.

\subsection{Tasks}

In the \textbf{topic-based task}, each participant answered fixed questions, with three questions per topic across six topics (see Table \ref{tab:task_questions}). Initially, participants identified the answers to the clinical questions themselves using the PDF guidelines from UCLH intranet. Then they inputted the questions into the chatbot to obtain its responses. The time was taken for both methods by the experiment observer.

Participants manually scored the chatbot answers against the answers they found on their own using two criteria: correctness (categories assigned: very relevant, relevant, irrelevant to the question) and completeness (categories assigned: minor omission, major omission, just right, minor addition, major addition).

In the \textbf{scenario-based task}, each participant was randomly assigned one scenario from a set of six predefined scenarios (see Table \ref{tab:scenarios}). Participants considered the scenario as if it were a real patient situation and were asked to report what information they would require from a clinical guideline to aid their diagnostic and treatment planning process. They were free to ask the chatbot three questions of their choice about the guideline. They searched for answers in the PDF guideline themselves and then via the chatbot. The time taken for both methods was measured by the observer.  

\paragraph{Data }

Following identification of the six clinical guidelines to be used, the pre-processing phase involved several steps. First, the guidelines were extracted from PDF documents using optical character recognition (OCR) technology to convert the text into a machine-readable format. This was followed by a manual cleaning process to correct any OCR errors and ensure the accuracy of the extracted text. Additionally, section names from each guideline were prepended to every line to facilitate the identification of pertinent information during the LLM analysis. Our guidelines have 47 lines and 950 words on average.

\section{Results }

\begin{table*}[h]
    \centering
    \begin{tabular}{lccc}
        \toprule
        \textbf{Category} & \textbf{Precision} & \textbf{Recall} & \textbf{F-score} \\
        \midrule
        Diagnosis: Hot Swollen Joint  & 0.10 & 1.00  & 0.18 \\
        Diagnosis: DKA               & 0.44 & 1.00 & 0.62 \\
        Diagnosis: Severe Hypertension & 0.28 & 1.00 & 0.44 \\
        Diagnosis: Pericarditis        & 0.85 & 1.00 & 0.92\\
        Diagnosis: Hypokalaemia        & 0.41 & 1.00 & 0.58 \\
        Diagnosis: Pancreatitis        & 0.51 & 1.00 & 0.68\\
        \midrule
        Average                & 0.43 & 1.00 & 0.57 \\
        \bottomrule
    \end{tabular}
    \caption{\textbf{Performance Evaluation of Guideline Chatbot Q\&A.} This table presents the precision, recall, and F-score based on the automatic evaluation of our chatbot answers per topic. The high recall values (1.00) across all diagnostic categories ensure that critical clinical information is not missed, which is essential for patient safety. While precision varies across categories, its impact may be less critical in this setting, as clinical experts can further filter and refine the retrieved information.}
    \label{tab:auto_eval}
\end{table*}

\paragraph{\textbf{Automatic Evaluation}}

The results in Table \ref{tab:auto_eval} show the precision, recall, and F-score across topics, based on an automatic evaluation of topic-based questions and golden answers as provided by FJ and NJ. The recall values remained consistently high (1.00) across topics, ensuring that critical clinical information is not missed. Our clinical experts conducted a thorough manual review of the chatbot outputs and confirmed that any no critical information for managing patient conditions was missed.

However, precision varies significantly, with Hot Swollen Joint (0.10) and Severe Hypertension (0.28) showing the lowest values. Pericarditis (0.85) indicate the strongest model performance in terms of precision. Note that precision is less critical in our setting, as clinical experts can further filter and refine the retrieved information (see examples in Table \ref{tab:examples}).

\paragraph{Human Evaluation} 

\begin{figure*}[ht]
\centering
\includegraphics[width=.7\linewidth]{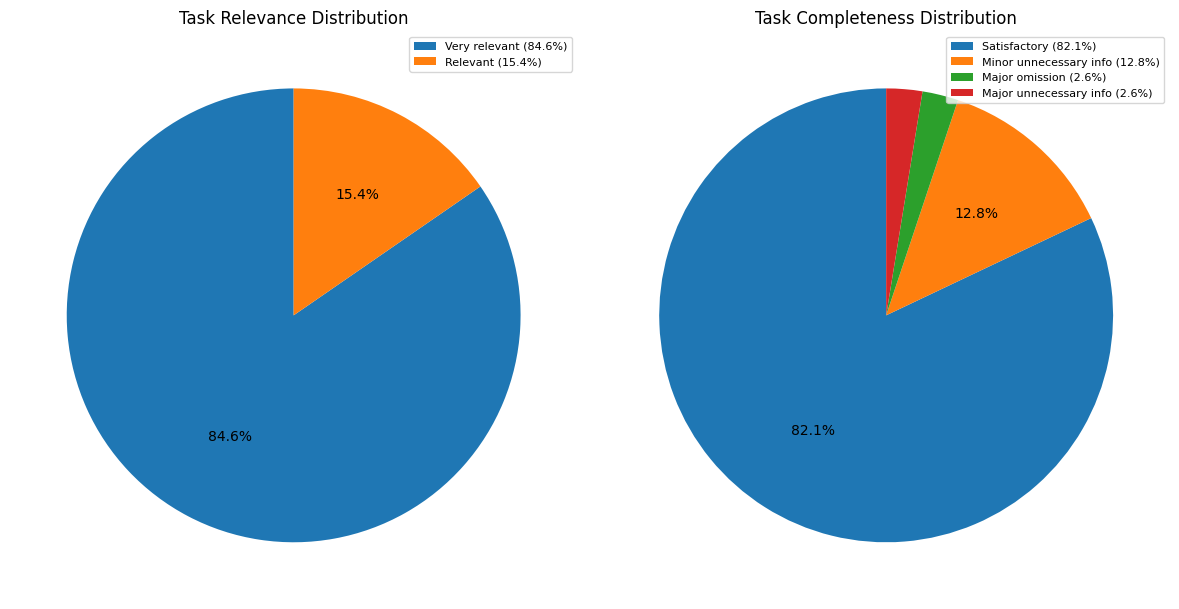}
\caption{\textbf{Task Relevance and Completeness Distribution Across Tasks.} Relevance distribution (left chart) shows that the majority of responses were deemed ``Very relevant'' (84.6\%). Completeness distribution (right chart) categorises responses based on the presence of unnecessary or missing information. A substantial majority (82.1\%) of responses were marked as ``Satisfactory''.} 
\label{fig:tasks}
\end{figure*}

In line with our previous findings, in the human evaluation setup our chatbot consistently extracts relevant information with high recall. During the topic-based task, when compared to manually identified answers, the chatbot produced ``very relevant'' responses 84.6\% of the time, and ``relevant information'' in 15.4\% of responses, with no ``irrelevant'' responses. Regarding the completeness of the chatbot answers, its responses were evaluated as being ``satisfactory'' in 82.1\% of cases, with minor additions of information occurring in 12.8\%  of responses. It is important to emphasise that our chatbot performs with minimal errors. Out of 39 response evaluations across various topics, there was only one instance each of major omissions and major unnecessary information (see Figure~\ref{fig:tasks}). Due to the hospital's connection quality, each LLM request was executed only once (compared to five times during automatic evaluation). We anticipate that consolidating multiple responses will further reduce the number of major errors (see Discussion section regarding deployment solutions).

In the scenario-based task, in which participants were able to ask their own freehand questions in response to a clinical scenario, the chatbot was evaluated as producing ``very relevant'' or ``relevant'' answers in 43.8\% and 50.0\% of cases, respectively. The chatbot gave only one ``irrelevant'' response. 

\begin{figure*}[ht]
\centering
\includegraphics[width=.7\linewidth]{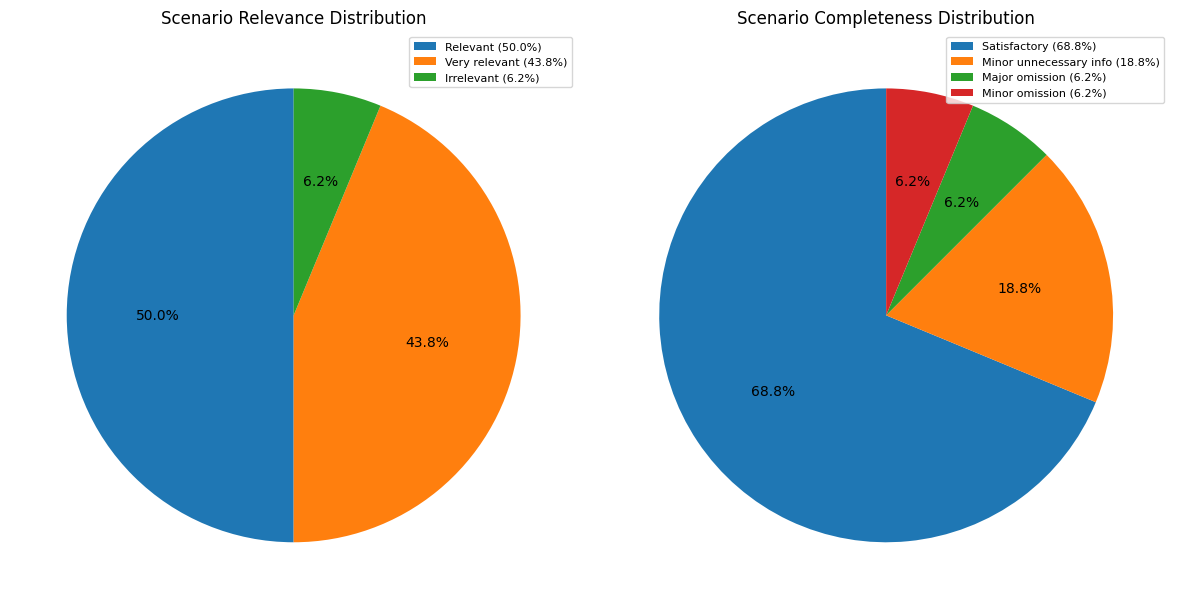}
\caption{\textbf{Task Relevance and Completeness Distribution Across Scenarios.} Relevance distribution (left chart) shows that the majority of responses were deemed ``Relevant'' (50.0\%). Completeness Distribution (right chart) categorises responses based on the presence of unnecessary or missing information. A substantial portion (68.8\%) of responses were marked as ``Satisfactory''. } 
\label{fig:scenario}
\end{figure*}

The completeness of these answers was evaluated as being ``satisfactory'' 68.8\% of the time, with a minor addition of information in 18.8\% of answers (see example of this information in~\ref{tab:example_auto}). There was only one response with minor omission and one response with major omission (see Figure~\ref{fig:scenario}). 

These findings and our observations during experiment highlight the importance of educating clinical experts on how to formulate questions that are closely aligned with the guidelines. Properly structured questions can significantly enhance the relevance and completeness of the chatbot's responses. It is important to note that that variations in doctors' questions has a moderate impact on the quality of the answers. In the worst-case scenario, if a question is vague or broad, the chatbot may provide a more extensive portion of the guideline in its output. This ensures that even less specific inquiries receive precise answers.

Moreover, our results suggest the promising advantage of our chatbot in terms of the time efficiency it offers.  All six participants working at NHNN reported needing to access the UCLH clinical guidelines either daily (n=2), or more than once per week (n=4), and four assessors stated they found it difficult or time consuming to access and navigate the clinical guidelines via the hospital intranet.

During the topic-based task, the mean time for assessors to identify answers to three fixed questions in the relevant guideline, including time navigating the UCLH intranet to find the guideline, was 90.15 seconds (range 57 - 178 seconds). The mean chatbot response time to answer the same questions was 34.42 seconds (range 27.47 - 40.14 seconds), including both the typing time from the participant, and the chatbot response time (see Table \ref{tab:timings}).

Finally, providing just line numbers to human experts was insufficient; they expressed need to understand the rationale behind the model's decisions. To address this, we explored the reasoning capabilities of Llama by examining the outputs of its DeepSeek version, \texttt{DeepSeek-R1-Distill-Llama-8B} \cite{deepseekai2025reasoningcapability}. Two assessors conducted this evaluation in parallel to their task-based and scenario-based evaluations (18 response evaluations in total). There was ``very good'' clinical reasoning found by the assessors in 72\% of the responses to the questions, with ``minor flaws'' found in 22\% of the time (see example in Table~\ref{tab:reasoning}). Only one response was marked with ``major flaw''.

\begin{table*}[h]
    \centering
    \begin{tabular}{lcccc}
        \toprule
        \textbf{Category} & \textbf{Total Count} & \textbf{Task Count} & \textbf{Scenario Count} & \textbf{Annotator} \\
        \midrule
        Diagnosis: Hot Swollen Joint  & 4 & 3 & 1 & R1, R2, R3, R4\\
        Diagnosis: DKA                & 4 & 3 & 1 & R1, R2, R6, R7\\
        Diagnosis: Severe Hypertension & 2 & 1 & 1 & R5, R6\\
        Diagnosis: Pericarditis        & 3 & 3 & 0 & R1, R3, R4\\
        Diagnosis: Hypokalaemia        & 3 & 2 & 1 & R2, R5, R6\\
        Diagnosis: Pancreatitis        & 3 & 1 & 2 & R3, R4, R5\\
        \midrule
        Total                & 19 & 13 & 6 \\
        \bottomrule
    \end{tabular}
    \caption{\textbf{Summary of Answers by Topic.} We report the total number of responses, the distribution between task-based and scenario-based answers, and the annotators involved.}
    \label{tab:summary_answers}
\end{table*}

\begin{table*}[h]
    \centering
    \begin{tabular}{l|p{5cm}|p{5cm}} 
        \toprule
        \textbf{Guideline} & \textbf{Mean participant response time per question, sec} & \textbf{Mean chatbot response time per question, sec} \\
        \midrule
        Hot Swollen Joint  & 88.33 &  40.14\\
        DKA                &  98.67 & 28.27 \\
        Severe Hypertension & 178.00 & 38.24 \\
        Pericarditis        & 57.00 & 35.84 \\
        Hypokalaemia        & 87.50 & 27.47 \\
        Pancreatitis        & 87.00 & 36.57 \\
        \midrule
        Total                & 90.15 & 34.42 \\
        \midrule
        \bottomrule
    \end{tabular}
    \caption{\textbf{Comparison of Response Times Between Participants and Chatbot for Topic-Based Tasks.} This table compares the mean response time (in seconds) for participants and the chatbot (including participant typing time) when answering three clinical questions related to different guidelines. The chatbot’s response time remains relatively stable across different conditions, averaging 34.42 seconds overall, whereas participant response times vary, with an overall mean of 90.15 seconds. }
    \label{tab:timings}
\end{table*}

\section{Discussion}

\paragraph{Statement of key findings}

Our results have demonstrated that concise and effective clinical information relevant to specific patient scenarios can be delivered by our LLM-based chatbot in response to clinical queries. The outputs were always clinically relevant, with an perfect 100\% recall. Furthermore, the overall speed of information retrieval for the clinician was threefold faster, including typing time, when using the chatbot compared to finding the information manually in the guideline.

Across all testing, there was no extemporised content or information given from outside the clinical guideline queried by the user. This is an important finding because of the potential safety implications in clinical practice if wrong drugs or doses are given, for example. By limiting the chatbot to only respond with verbatim lines copied from the clinical guideline we were able to ensure it maintained clinically safe responses. Thus, although there is a reasoning approach by the model to assess the appropriate output in response to a question by the user, this does not involve text generation, but merely the choice of already existing text in the clinical guideline.

Overall, there was a slightly lower performance for the chatbot when asked ``freehand'' questions by clinicians relating to a clinical scenario. This is likely due to the phrasing of prompt questions by clinicians not accustomed to interacting with LLM tools, plus early stage inflexibility of the model, and will improve with wider implementation and clinician experience as well as model updates. There may be value in providing clinicians with a ``primer'' on effective ways to interact with the model to optimise efficiency of output.

\paragraph{Strengths and Limitations}

This study uses a novel approach to successfully constrain an LLM model in order to allow clinically relevant, safe, disease-specific information in response to a targeted clinical question. Additionally, we assessed the reasoning capacities of the model, which represents an understudied area. Although using clinicians to assess the output relevant to their daily workload provides the study with real world value, the sample size used was small and larger studies would likely provide more insight into how this model could be improved. 

The guidelines used at UCLH are text-based and contain no flowcharts or visual aids. However, it is possible that future iterations of guidelines or those used at other hospitals could contain these features and so it is important to augment the chatbot with capability to convert visual flowcharts to text for future iterations. 

Finally, we used simulated patient scenarios as the basis for interaction with the model. Deploying this in the real world clinical setting with clinicians using the model when required in real patient situations would be the optimal way to assess its utility in future studies. The LLM we use is one of the smallest available, allowing it to be hosted locally on a server that can be easily installed within the hospital environment. This setup does not require costly computing hardware and infrastructure.

\paragraph{Implications and next steps}

This pilot feasibility study has demonstrated that an LLM model can effectively assist clinicians managing patient issues by sourcing relevant and useful information from clinical guidelines. With increasing time pressures experienced by clinicians across specialities, this time saving is a potentially powerful patient safety tool as it also ensures up to date information is available to the clinician. 

We plan to conduct a comprehensive evaluation of the chatbot’s accessibility and usability among healthcare professionals. In addition to the parameters assessed in this pilot study, it is essential to evaluate its performance and acceptability across diverse user groups, including those from lower socioeconomic backgrounds \cite{patel2022ethnic,rhead2024ethnic}. Such an evaluation is critical to ensure the tool contributes to reducing, rather than exacerbating, the digital divide—an important consideration in equitable healthcare delivery.

Further iterations of this tool should be tested with real-world scenarios in different healthcare settings (e.g. primary care, emergency medicine, hospital care). Additionally, we plan to extend the application to real-life settings for use over the phone, expand its reference materials to include more guidelines, phonebooks, and antibiotic manuals, and enhance its decision tree reasoning capabilities. Equipping the chatbot with the ability to ask context-specific questions to refine the clinical query will further improve its utility.  

As many hospital trusts around the UK are adopting electronic health record (EHR) software for clinical management of inpatients, our tool could ideally be integrated with these systems which would allow further added functionality by giving the chatbot access to patient-specific information which can shape the response (e.g. advising the right dose of intravenous antibiotic based on patient's recorded weight). However, for hospitals not yet using EHR systems, the development of a clinical app will be required to allow widespread use of our chatbot, accessible by staff using hospital trust credentials to ensure clinical safety.

\subsection{Conclusion}

The adherence to clinical guidelines in primary and secondary care improves patient outcomes by providing evidence-based standards of care. This feasibility study demonstrates that implementation of an AI-based helper tool, ideally integrated with patient electronic health records, to search for and summarise key clinical recommendations from these guidelines, has the potential to optimise guideline adherence, improve patient safety, and increase patient-facing time for clinicians. Moreover, the information extraction capacities of Large Language Models (LLMs) open new perspectives for safe and transparent AI support, enhancing the ability to deliver contextually relevant clinical information.

\section*{Code and Data Availability}

The code and data could not be publicly shared due to their confidentiality requirement.

\section*{Competing interests} We have no competing interests to declare. 

\section*{Author Contributions Statement}

JI: Conceptualisation, Methodology, Software, Validation, Formal analysis, Investigation, Writing – Final draft preparation, Writing - Reviewing and Editing. FJ, NJ: Conceptualisation, Methodology, Formal Analysis, Investigation, Writing – Early draft preparation, Writing - Reviewing and Editing. CSH, PB: Writing - Reviewing and Editing. RD: Conceptualisation, Methodology, Writing - Reviewing and Editing. All authors approved the manuscript. 

\bibliographystyle{aaai}
\bibliography{lit}

\section*{Appendices}

\setcounter{table}{0}
\renewcommand{\thetable}{A\arabic{table}}

\begin{table*}[h]
    \centering
    \scalebox{0.9}{
    \begin{tabular}{p{2cm}|p{10cm}}
    \toprule
    \textbf{Topic} & \textbf{Questions} \\
        \toprule
         Hot Swollen Joint & What differentials are there for a hot, swollen joint? \newline What investigations should I do for septic arthritis? \newline
         What is first line treatment for a gout flare?  \\
        \midrule
       Pericarditis & What blood tests should I request for a patient with pericarditis? \newline
What is the first-line treatment for pericarditis? \newline
Do I need to admit a patient if they have new pericarditis? \\
\midrule
 Diabetic ketoacidosis (DKA) & What are the criteria for diagnosing DKA? \newline
What is the fluid management in DKA? \newline
How do you define severe DKA?  \\
        \midrule
Hypokalaemia & How do we treat hypokalemia?  \newline
What does an ECG show in hypokalemia? \newline
What monitoring is required in hypokalemia? \\
        \midrule
Severe Hypertension & What investigations should I request for a patient with severe hypertension? \newline
How do I define hypertensive emergency? \newline
What dose of GTN should I administer IV for hypertensive emergency? \\
        \midrule
Pancreatitis & What investigations should I request for a patient with pancreatitis? \newline
Who can I contact for advice about my patient with pancreatitis? \newline 
How should I manage a patient with pancreatitis?  \\
        \bottomrule
    \end{tabular}}
    \caption{\textbf{Task Questions Categorised by Clinical Topics for Human Evaluation.} The topics cover a range of critical diagnoses. The questions focus on key aspects of clinical decision-making, such as diagnosis criteria, investigations, treatment options, and patient management strategies. It is important to note that ``Golden Answers'' (expert-validated responses) are not provided due to the proprietary nature of the data. }
    \label{tab:task_questions}
\end{table*}

\begin{table*}[h]
    \centering
    \scalebox{0.9}{
    \begin{tabular}{p{2cm}p{16cm}}
    \toprule
    \textbf{Topic} & \textbf{Scenario} \\
        \toprule
         Hot Swollen Joint & 60 M post-op D5 meningioma resection, complaining of tender swollen right knee. O/E - right knee hot to touch, ROM limited by pain, NVI distally, calf soft. Obs – T 37.5, HR 90, BP 120/81.  \\
        \midrule
       Pericarditis &  40 M, post-op D8 tumour biopsy, due to go home today. New left sided chest pain, alleviated by sitting forward. T 37.5, HR 90, BP 120/90. ECG shows saddle shaped ST elevation. Bloods – troponin 3. Recent coryzal symptoms. \\
\midrule
 Diabetic ketoacidosis (DKA) &   21 F with PMH T1DM, admitted today for ICP monitoring ?IIH. Medications not yet charted. On ward round, CBG 24 with ketones 3.0. VBG result: pH 7.3. Obs normal.  \\
        \midrule
Hypokalaemia &  30 M post-op D1 Vp shunt. Nausea and vomiting ++. Post-op bloods: Potassium 2.3 (3.5-5.5), Sodium 137 (135-145), eGFR >90, Ca 2.2 (2.2-2.6). \\
        \midrule
Severe Hypertension &  68 M admitted for elective brain biopsy. Pre-op observations - BP 190/110 confirmed 2 separate readings. HR 80, T 36.3. Bloods – Na 135 (135-145), K 3.8 (3.5-5), Cr 138 (60-110), eGFR 70 (>90). No history of kidney disease.  \\
        \midrule
Pancreatitis & 50 F, post-op day 3 L4/5 discectomy. PMH gallstones. Post-op abdominal pain, epigastric radiating to back. Obs: sats 100\% on RA, HR 110, BP 115/80, T 36.5. Bloods show amylase 700 (30-118), bilirubin 3, CRP 50, WCC 6 (normal neutrophils), Hb 110, Na 135 (135-145), K 3.8 (3.5-5).  \\
        \bottomrule
    \end{tabular}}
    \caption{\textbf{Clinical Scenarios Categorised by Topic for Human Evaluation.} Each scenario describes a patient case, including relevant history, symptoms, observations, and lab results, to simulate real-world decision-making.
    \label{tab:scenarios}}
\end{table*}

\begin{table*}[h]
    \centering
    \scalebox{0.9}{
    \begin{tabular}{p{16cm}}
        \toprule
          \textbf{Topic}: Pericarditis \\
\textbf{Question}:Do I need to admit a patient if they have new pericarditis? \\
\textbf{Chatbot Response}:  
DECISION TO ADMIT FROM THE ED \newline

Admit if haemodynamically unstable \newline

Otherwise admit if: \newline

    \end{tabular}}
    \caption{\textbf{Examples of our Ground Truth Response related to Pericarditis}. The bolded text highlights phrases that were added in the chatbot response during automatic evaluation (ensemble chatbot response from five runs). The human experts who created the ``golden responses'' acknowledged the supportive rather than essential role of the highlighted phrases. All the examples have been paraphrased.}
    \label{tab:example_auto}
 \end{table*}

\begin{table*}[h]
    \centering
    \scalebox{0.9}{
    \begin{tabular}{p{16cm}}
        \toprule
          \textbf{Topic}: Severe Hypertension \newline
\textbf{Question}: What dose of GTN should I administer IV / intravenous for hypertensive emergency? \newline
\textbf{Response}: \newline Please follow the steps below: \newline

Blood Pressure Targets \newline
Unless there is an aortic dissection, the following targets apply: \newline

First Hour Using IV Drugs: \newline
Reduce Mean Arterial Pressure (MAP) by a maximum of 20\%. \newline

2-6 Hours Using IV Drugs: \newline
Aim for a blood pressure of 160/100 mmHg. \newline
If blood pressure remains above target after the maximum dose of a single IV agent, start a second IV agent. \newline

After the First 6 Hours: \newline
Begin introducing oral agents and gradually reduce IV drugs. \newline
Aim for a blood pressure of 140/80 mmHg over the next 24-48 hours. \newline

\textbf{Antihypertensives }\newline
\textbf{Intravenous Glyceryl Trinitrate (GTN):} \newline

\textbf{Prepare 50 mg GTN in a total of 50 ml of 0.9\% sodium chloride.} \newline
\textbf{Infuse at a rate of 1-10 mg/hour, with higher rates at the consultant's discretion.} \newline
\textbf{Recommended starting rate is 1-3 mg/hour.} \newline
\textbf{Adjust the infusion rate in increments of 1-2 mg/hour every 15 minutes.} \newline
\textbf{Note: GTN becomes less effective over time due to tolerance, so avoid use for more than 24 hours.} \newline

Please refer to Severe Hypertension guidelines for more information. \\
\midrule
\textbf{Topic}: Pancreatitis \newline
\textbf{Question}: What investigations should I request for a patient with pancreatitis? \newline
\textbf{Response}: \newline
Please follow the steps below: \newline

\textbf{Investigations} \newline

\textbf{Blood Tests:} \newline
\textbf{Venous Blood Gas (VBG)} \newline
\textbf{Full Blood Count (FBC)}\newline
\textbf{Urea and Electrolytes (U\&E) } \newline
\textbf{Liver Function Tests (LFT)}\newline
\textbf{Amylase }\newline
\textbf{C-Reactive Protein (CRP)}\newline
\textbf{Clotting profile }\newline
\textbf{Bone profile }\newline
\textbf{Triglycerides} \newline
\textbf{Blood cultures (if indicated)} \newline

\textbf{Additional Tests:}\newline
\textbf{Consider Arterial Blood Gas (ABG) to assess hypoxia if present.} \newline
\textbf{Pregnancy test if appropriate. }\newline

\textbf{Imaging: }\newline
\textbf{Chest X-ray (CXR) to assess for effusion or Acute Respiratory Distress Syndrome (ARDS). }\newline
\textbf{Abdominal Ultrasound (USS) to determine the underlying cause. }\newline
\textbf{Computed Tomography (CT) scan only if there is peritonitis or diagnostic uncertainty.}\newline

Please refer to Pancreatitis guidelines for more information.
\\
        \bottomrule
    \end{tabular}}
    \caption{\textbf{Examples of our Chatbot Responses related to Severe Hypertension and Pancreatitis}. Each example consists of a question followed by a structured response. The bolded text within the responses highlights phrases that match the ``golden answer'' provided by human experts emphasising the high recall of our chatbot answers. All the examples have been paraphrased.}
    \label{tab:examples}
\end{table*}

\begin{table*}[h]
    \centering
    \scalebox{0.9}{
    \begin{tabular}{p{16cm}}
    \toprule
\textbf{Question from the clinician:} How do I diagnose DKA? \newline

\textbf{Output from chatbot:}   \newline 

Okay, so I need to figure out which lines in the provided context are highly relevant to diagnosing DKA. The question is asking for the criteria, so I should focus on lines that outline the specific diagnostic markers.

Looking at line 1, it clearly lists the three criteria: blood glucose, blood ketones, and acidosis. That's directly about the diagnosis, so that's definitely relevant.

Line 0 mentions performing CBG, VBG, and urinalysis, which are the tests needed to assess blood glucose and ketones, so that's also relevant.

Line 2 talks about considering causes of DKA, which is more about differential diagnosis rather than the specific criteria, so maybe not as relevant.

Lines 3-6 are about requests for tests and observations, which are part of the diagnostic process but don't specify the criteria themselves.

Lines 7-9 discuss fluid management, which is important but not directly about the diagnostic criteria.

Lines 10-16 are about potassium management, which is part of treatment but not the diagnosis itself.

Lines 17-18 deal with insulin management, which is part of treatment, not diagnosis.

Lines 19-21 talk about severe DKA and prophylaxis, which are more about management and prevention rather than diagnosis criteria.

Lines 22-26 discuss monitoring and targets, which are important but again, more about management than initial diagnosis.

Lines 27-30 are about resolution and referrals, which are post-diagnostic steps.

So, the most relevant lines are line 0 and line 1 because they directly address the diagnostic tests and criteria needed for DKA. \newline

$<$/think$>$ \newline

[0 1] \newline
ANSWER TO THE USER: \newline

DIAGNOSIS \newline
PERFORM A CBG, VBG AND URINALYSIS \newline
Blood glucose levels of at least 11.1 mmol/L or a known history of diabetes. \newline
Blood ketones of 3 mmol/L or higher, or urine ketones greater than 2+. \newline
Acidosis, indicated by venous or arterial pH below 7.3 and/or HCO3- levels under 15 mmol/L.\newline

Please refer to DKA guidelines for more information.\\
        \bottomrule
    \end{tabular}}
    \caption{\textbf{Chatbot's Line Selection Process with 
DeepSeek-R1-Distill-Llama.} We show our chatbot's reasoning process when determining the most relevant lines from the provided guideline. All the guideline text has been paraphrased.
    \label{tab:reasoning}}
\end{table*}

\end{document}